\documentclass{article}

\usepackage[final]{neurips}

\makeatletter
\renewcommand{\@noticestring}{}
\makeatother

\usepackage{float}
\usepackage{needspace}

\makeatletter
\renewenvironment{table}
  {\setlength{\abovecaptionskip}{\baselineskip}%
   \setlength{\belowcaptionskip}{0pt}%
   \@float{table}}
  {\end@float}
\makeatother

\usepackage[utf8]{inputenc}
\usepackage[T1]{fontenc}
\usepackage{amsmath,amssymb}
\usepackage{amsthm}
\usepackage{graphicx}
\usepackage{booktabs}
\usepackage{xcolor}
\usepackage[hidelinks]{hyperref}
\usepackage{url}

\graphicspath{{figures/}}
\theoremstyle{plain}

\newcommand{\R}{\mathbb{R}}
\newcommand{\rank}{\operatorname{rank}}
\newcommand{\recninety}{\mathrm{rec}@0.9}
\newcommand{\Kmat}{K_{\mathrm{mat}}}
\newcommand{\Vmat}{V_{\mathrm{mat}}}

\title{When the Gradient Sees Rank: Provable
Necessity, Causal Recruitment, and Composition in Trained Matrix
Memories}

\author{%
  Samuel Larson\\
  Pebble ML\\
  \texttt{samlarson@pebbleml.com}%
}

\begin{document}

\maketitle

\begin{abstract}
Can gradient-based training learn the rank needed to store and compose
associations in a matrix memory? In our earlier study, we used a
matrix-augmented reasoner on a task that admits a rank-1 solution,
leaving this question open. We train matrix memories on $K$ fresh key--value
bindings whose exact linear recovery requires $\rank(Z) \geq K$.
A fixed linear readout queries a single matrix state without access to
the original bindings. Experiments measure recovery by cosine similarity
greater than 0.9, a threshold distinct from mathematical equality.
Learned effective rank increases with $K$ across the tested grid
(Spearman $\rho = 1.0$ at $d = 16$).
Training-time rank caps produce a recovery transition near $k = K$:
at $d = 8$, $K = 4$, rank 3 gives at most 0.0004 recovery and rank 4
gives 0.97.
Four of five seeds retain at least 0.9996 recovery through 21-fold
self-application of the trained operator.
On the entity subspace, the learned operator has effective rank close
to $K$ and approximates the ideal cycle.
For the single converged seed capped below $K$, a calculation using
the entity-subspace operator and ideal cycle predicts the measured
cosine within 0.008 through seven applications.
Extending training resolves several initial failures, but recovery still
declines at larger matrix dimensions with encoder width fixed.
\end{abstract}


\section{Introduction}
\label{sec:intro}

Matrix-valued states appear in fast-weight architectures
\citep{schlag2021linear} and extensions of continuous chain-of-thought
models \citep{hao2024coconut}. Their rank describes the dimension of the
image of the linear transformation they represent. Does training learn
the rank needed for a task, and does restricting that rank impair
performance?

A recent study added a $d \times d$ matrix latent to a vector-pretrained
GPT-2 reasoner and found flat rank-$k$ truncation curves across four
training regimes and four readouts \citep{larson2026gradient}. The study
identified a limitation: its ProsQA task admits a rank-1 solution.
That result therefore does not establish how training behaves when
exact recovery requires higher rank.

We study models trained from scratch on tasks whose exact linear
solutions require $\rank(Z) \geq K$. A fixed readout and a single matrix
state isolate the role of rank (\S\ref{sec:setup}). We measure the rank
learned during training and the effect of training-time rank caps
(\S\ref{sec:recruit}). We then test repeated application of the learned
operator, analyze its action on the entity subspace
(\S\ref{sec:compose}), and examine recovery at larger matrix dimensions
and longer training budgets (\S\ref{sec:frontier}).

\section{Task, Rank Requirement, and Experimental Controls}
\label{sec:setup}

\textbf{Task and rank requirement.} Each episode presents $K$
key--value pairs, then queries one key. Keys and values are sampled
anew for each episode (Gaussian, normalized, or exactly orthonormal
following \citealp{mezzadri2007random}), so evaluation uses new
bindings. Write the keys and values as the columns of $\Kmat$ and
$\Vmat$. Exact recovery, $Z k_j = v_j$ for all $K$ bindings with
linearly independent keys and values, requires $Z \Kmat = \Vmat$.
The classical rank inequality then gives
\[
K = \rank(\Vmat) = \rank(Z \Kmat) \leq \rank(Z)
\]
\citep{kohonen1972correlation,anderson1972simple}.
We test whether training with Adam learns the required rank and how
restricting rank affects recovery.

\textbf{Fixed linear readout.}
\citet{nichani2025factual} construct a rank-$m$ linear memory storing
approximately $md$ associations under interference-tolerant argmax
decoding. The exact-recovery rank requirement above does not apply to
that decoding rule. Here, the prediction is the vector $Z k_j$, with
no argmax or learned readout. Training minimizes one minus cosine
similarity. We report $\recninety$, the fraction of queries with
cosine similarity greater than 0.9. This threshold measures approximate
recovery; it does not require $Z k_j = v_j$ exactly.

\textbf{Single-state bottleneck.}
A full-attention decoder could retrieve individual bindings from
separate input positions. To prevent this, the decoder reads only
$(Z, \mathrm{query})$, where $Z$ is one $d \times d$ state. After the
encoder writes $Z$, we corrupt the raw-input cache and confirm that
the decoder output is bit-for-bit unchanged.

\textbf{Model and rank controls.} A Transformer encoder processes the
binding tokens. It uses $d$ learned row-reader latents to produce the
rows of $Z \in \R^{d \times d}$. The mapping is permutation-invariant,
can express rank up to $d$, and does not hard-code
$\sum_j v_j k_j^\top$. Every model has fewer than 1M parameters; the
$d = 16$ composition model has 170{,}896.
Effective rank, the exponential of the entropy of the normalized
singular-value spectrum, is the pre-registered primary metric.
In the causal control, spectral projection restricts $Z$ to rank at
most $k$ at every training step, with backpropagation through the
projection. The restriction is therefore part of training, rather
than an intervention applied only after training.

\section{Learned Rank and the Effect of Rank Caps}
\label{sec:recruit}

\textbf{Learned rank.} Without a rank penalty in the loss, effective
rank increases monotonically with binding count on the ten-point
$d{=}16$ grid from $K{=}1$ to $K{=}16$ (Spearman $\rho = 1.0$).
It is 2.42 at $K{=}1$, 8.20 at $K{=}8$, and 15.09 at $K{=}16$.
It exceeds the pre-registered $[0.7K, 1.3K]$ band at $K \leq 3$
and lies within the band at the other tested points. Beyond $K = d$,
effective rank saturates and then declines.
A subsequent, larger replication sweep shows the same overall pattern.

\begin{figure}[htbp]
\centering
\includegraphics[width=\textwidth]{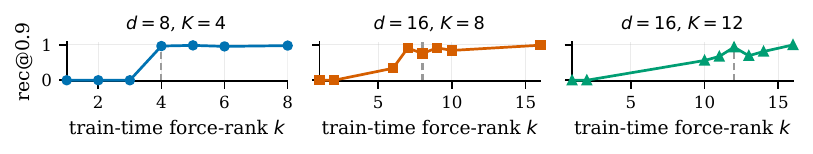}
\caption{Recovery under training-time rank caps in the pre-registered
causal test. $\recninety$ is the fraction of queries with cosine
similarity greater than 0.9; the dashed line marks $k = K$.
The archived sweep shows recovery increasing near $k = K$, with a
more gradual transition for $d{=}16$, $K{=}12$.}
\label{fig:forcerank}
\end{figure}

\textbf{Training-time rank caps.} At $d{=}8, K{=}4$, training with
rank cap $k \leq 3$ yields $\recninety \leq 0.0004$, while $k = 4$
yields 0.97 (Figure~\ref{fig:forcerank}). The observed transition
occurs at the rank required for exact recovery.
The $d{=}16$ configurations also show a transition near $k = K$,
with some non-monotonic variation above it. In the replication, the
$K{=}12$ transition is a gradual, irregular rise rather than a step.
These results show that gradient-based training can learn and use
higher rank in this setting. The flat curves in
\citet{larson2026gradient} do not imply a general inability to do so.

\section{Composition and the Entity Subspace}
\label{sec:compose}

The composition task arranges $K$ entities in one cycle $\pi$.
The entities are represented by orthonormal vectors $e_1,\ldots,e_K$.
The one-step binding
is $k_i=e_i$, $v_i=e_{\pi(i)}$, so the output uses the same vector
space as the next input. A query at depth $h$ asks for $e_{\pi^h(i)}$,
obtained by applying the same trained operator $h$ times: $Z^h e_i$.
There are no learned
per-hop parameters. Training uses depths $\{1, 2, 3\}$; evaluation
adds $\{4, 5, 6\}$ and probes $\{7, 21\}$ ($d = 16$, $K = 8$,
orthonormal keys). A single full cycle avoids the shorter cycles of
a general random permutation, which can turn nominally held-out
depths into identity or training queries. Depth 21 has the same
target as depth 5 in an 8-cycle; it tests stability under additional
applications (Appendix~\ref{app:period}).

\textbf{Recovery across depths.} Four of five unconstrained seeds
reach $\recninety \geq 0.9996$ at every tested depth; three reach
1.000 throughout. At the training budget, the fifth seed reaches
0.93 at $h{=}1$, falling to 0.16 at $h{=}7$ and 0.0001 at $h{=}21$.

\begin{figure}[htbp]
\centering
\includegraphics[width=\textwidth]{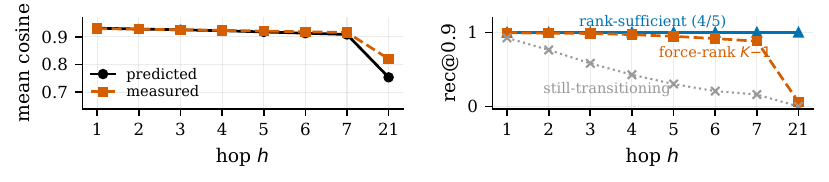}
\caption{Recovery under repeated application of the learned matrix
($d{=}16$, $K{=}8$). \emph{Left:} mean cosine for the converged seed
with rank cap $k{=}7{=}K{-}1$, compared with a prediction computed
from its entity-subspace operator and the ideal cycle. \emph{Right:} the fraction of queries with
cosine greater than 0.9, averaged over the four converged
unconstrained seeds (near 1.0), and shown separately for the
rank-capped seed and the fifth unconstrained seed.}
\label{fig:depth}
\end{figure}

\textbf{Error accumulation.} With rank cap $k = K - 1$, exact linear
recovery is impossible. Nevertheless, the one converged seed reaches
mean cosine 0.92--0.93 at shallow depths ($h \leq 5$).
Its $\recninety$ falls from 0.996 at $h = 1$ to 0.881 at $h = 7$
and 0.060 at $h = 21$ (Figure~\ref{fig:depth}).
A calculation using the entity-subspace operator and ideal cycle
predicts measured cosine within 0.008 through $h = 7$ (Table~\ref{tab:depthcurve}).

\textbf{Action on the entity subspace.} The converged solutions have
whole-matrix effective rank 16.0 at $d{=}16$ and stable rank
14.6--15.6. These whole-matrix measures include components outside
the subspace used by the task.
We therefore decompose $Z$ into blocks relative to the ideal cycle's
$K$-dimensional entity subspace $\mathcal{E}$. The block acting within
$\mathcal{E}$ has effective rank 7.999--8.000 (maximum 8) and a
scale-corrected residual of 0.7--2.4\% from the ideal cycle. Leakage
out of $\mathcal{E}$ is below 1.5\% of the state norm. The complement
block is full-rank and non-contractive
(Table~\ref{tab:subspace}, Appendix~\ref{app:subspace}).
These measurements support an approximately invariant entity
subspace on which effective rank is close to $K$. Rank-capped seeds
that fail recovery also fail the corresponding subspace test.

\section{Training Budget and Recovery at Larger Dimensions}
\label{sec:frontier}

\textbf{Longer training.} Several configurations that failed at their
initial training budget succeeded with more steps. For composition
at $K{=}8$, one of five seeds converged at 20K steps and four of five
at 40K.
At $K \in \{12, 16\}$, none of five seeds converged at 40K; at 80K,
three of three and two of three fresh seeds converged, respectively.
In the dimension sweep, configurations at $d \geq 32$ failed at an
8K-step budget but began learning at 6--16K steps when training was
extended to 20K.
These reversals motivate re-testing unsuccessful configurations at
2--2.5$\times$ the initial budget before drawing conclusions about
trainability.
An extension does not guarantee recovery: the remaining unsuccessful
$K{=}16$ seed still failed at 120K steps,
and the $d{=}32$ configurations below plateaued below the
pre-registered success threshold.

\Needspace{9\baselineskip}
\textbf{Recovery at larger dimensions.} With $K = d/4$ and encoder
width fixed at 64, mean recovery cosine declines as $d$ increases.
All reported runs had plateaued: at $d{=}32$, three seeds reached
0.877/0.909/0.915 after 100K steps; at $d{=}48$, one seed reached
0.7196 after 100K; at $d{=}64$, one seed reached 0.3882 after 150K.
Their $\recninety$ scores were 0.413/0.632/0.653 at $d{=}32$
(all below the pre-registered 0.7 success threshold), 0.002 at
$d{=}48$, and 0.0 at $d{=}64$.
At $d{=}32$, learned effective rank reaches 91--97\% of the target,
but recovery remains below the success threshold. High effective rank
alone does not ensure accurate bindings.

\section{Related Work and Limitations}
\label{sec:related}

\citet{nichani2025factual} construct linear memories under
interference-tolerant argmax decoding, and \citet{barnfield2026sharp}
derive static-memory thresholds. Here, the readout is fixed to the
matrix--vector product, and we measure and restrict the rank learned
during training. \citet{nazari2026rank} and \citet{sun2026staterank}
study effective-rank dynamics in pretrained linear-attention models
observationally, including the upper bound $\rank(S_t) \leq t$.
Our experiments control the number of bindings and impose rank caps
during training.

\citet{mishra2026m2rnn} train matrix-state RNNs on $S_3$ without rank
measurement or intervention; their vector-state control matches
performance. \citet{grazzi2025negative} and
\citet{siems2025deltaproduct}, the latter extending
\citet{yang2024deltanet} with Householder products, characterize the
per-token transition operator. We instead study the rank of the
trained memory state. Continuous-CoT capacity results give existence
constructions \citep{zhu2025superposition,gozeten2025cot2}. Work on
compositional generalization
\citep{dziri2023faithfate,wang2024grokked} motivates examining the
learned operator as well as its recovery scores.

\textbf{Limitations.} Every experiment is synthetic and uses fewer
than 1M parameters. Generality at scale and across state-writing
architectures, including recurrent fast-weight, attention-readout,
and associative-memory models, remains untested.
The comparison between predicted cosine and recovery across depths
rests on one converged seed with rank cap $K{-}1$. The other two
seeds in that configuration failed under a documented numerical
instability in the spectral-projection backward pass. This is a
single-seed case study, and the $d \geq 48$ results also use single
seeds. Appendix~\ref{app:repro} lists the pre-registered criteria
for challenging the empirical conclusions.

A companion paper on group-composition state tracking
\citep{larson2026ranklaw} uses the binding setup by citation. It shares
no figures or tables with this paper and studies a group-dimension
rank law beyond the scope of the present experiments.

\bibliographystyle{plainnat}
\bibliography{refs}

\appendix

\section{Depth 21 Under the Single $K$-Cycle}
\label{app:period}

Under a single 8-cycle, $\pi^{21} = \pi^{21 \bmod 8} = \pi^{5}$.
Depth 21 therefore shares its target with depth 5; the two are not
distinct group-theoretic queries. The depth-21 probe measures
stability under 21 sequential applications of the trained operator.
In the four converged seeds, $\recninety \geq 0.9996$ at both depths,
so the additional applications preserve the high recovered fraction.
For the rank-capped operator, the recovered fraction falls from
0.947 at $h{=}5$ to 0.060 at $h{=}21$.
This comparison tests error accumulation, rather than generalization
to an unseen permutation target.

\begin{table}[H]
\centering
\small
\begin{tabular}{cccc}
\toprule
$h$ & predicted $\cos$ & measured $\overline{\cos}$ & measured $\recninety$ \\
\midrule
1  & 0.9317 & 0.9303 & 0.996 \\
3  & 0.9258 & 0.9259 & 0.986 \\
5  & 0.9181 & 0.9212 & 0.947 \\
7  & 0.9089 & 0.9163 & 0.881 \\
21 & 0.7536 & 0.8206 & 0.060 \\
\bottomrule
\end{tabular}
\caption{Recovery across depths for the operator with $d=16$, $K=8$,
and rank cap $k = 7 = K - 1$. Cosine predicted using the entity-subspace operator and ideal cycle
(without raw keys) is compared with measured mean cosine and the
fraction of queries with cosine greater than 0.9. Prediction error
is below 0.008 through $h = 7$. At $h = 21$, mean cosine remains
0.8206, but only 0.060 of queries exceed the recovery threshold.}
\label{tab:depthcurve}
\end{table}

\Needspace{24\baselineskip}
\section{Per-Seed Subspace Decomposition}
\label{app:subspace}

\begin{table}[H]
\centering
\small
\begin{tabular}{lccccc}
\toprule
seed & $\mathrm{effrank}(A)$ & residual & leakage (\% of $\|Z\|$) & $\mathrm{effrank}(D)$ & $\rho(D)$ \\
\midrule
s1 & 8.000 & 0.7\% & 0.44\% & 8.000 & 1.38 \\
s2 & 7.999 & 1.9\% & 1.06\% & 8.000 & 1.27 \\
s3 & 7.999 & 1.5\% & 0.87\% & 8.000 & 2.86 \\
s4 & 7.999 & 2.4\% & 1.45\% & 7.999 & 1.02 \\
\bottomrule
\end{tabular}
\caption{Entity-subspace decomposition of four converged $d{=}16$, $K{=}8$
composition seeds, recomputed from archived matrices and averaged
over four evaluation episodes. $A$ acts within the entity subspace.
Residual is the scale-corrected Frobenius distance to the ideal cycle,
after fitting the isotropic scale that cosine scoring does not
measure. Leakage is the cross-block norm $\|C\|$ as a fraction of
$\|Z\|$. $D$ acts within the complementary subspace;
$\rho(D)$ is its spectral radius. The complement is full-rank and
non-contractive, while the measured cross-block leakage is small
but nonzero.}
\label{tab:subspace}
\end{table}

\section{Reproducibility}
\label{app:repro}

Training, evaluation, and analysis code and archived results are
available in the public repository.\footnote{\url{https://github.com/saml212/matrix-memories}}
Per-run JSON archives support the reported numbers. A versioned script regenerates each figure
and verifies the checksum of every source file it loads. Hypotheses,
metrics, rank controls, and decision bands were registered in design
documents before training. One metric was re-registered before the
affected sweep; that deviation is recorded in those documents.

The pre-registered criteria for challenging the empirical conclusions
remain open: a rank cap of $K{-}1$ achieving at least
$0.9\times$ the converged unconstrained model's recovery at multiple
$(d, K)$ points, or a $d \geq 32$ configuration clearing the 0.7 recovery
threshold. These criteria concern measured recovery under the
experimental conditions, not the algebraic requirement for exact
linear recovery.

The rank-cap sweep aggregate retains one recovery value per
$(d, K, k)$ configuration, without per-seed instability diagnostics.
It therefore describes the observed recovery transition but cannot
separate representational limits from numerical training failures
in individual runs. Variation above the transition should likewise
be interpreted at the aggregate level.

On some platforms, the analysis pipeline emits BLAS
\texttt{RuntimeWarning}s for near-singular intermediate products.
Every decomposition output is checked for finiteness before use.

\end{document}